# Blind Deconvolution Method using Omnidirectional Gabor Filter-based Edge Information

Trung Dung Do, Xuenan Cui, Thi Hai Binh Nguyen, Hakil Kim
School of Information and Communication Engineering
Inha University, Incheon, Korea
dotd@inha.edu, xncui@ inha.ac.kr, binhnth12@gmail.com, hikim@inha.ac.kr

Van Huan Nguyen
Faculty of Information Technology, Ton Duc Thang University, Ho Chi Minh City, Vietnam.
nguyenvanhuan@tdt.edu.vn

*Abstract*— **In the previous blind deconvolution methods, de-blurred images can be obtained by using the edge or pixel information. However, the existing edge-based methods did not take the advantage of edge information in ommi-directions, but only used horizontal and vertical edges when recovering the de-blurred images. This limitation lowers the quality of the recovered images. This paper proposes a method which utilizes edges in different directions to recover the true sharp image. We also provide a statistical table score to show how many directions are enough to recover a high quality true sharp image. In order to grade the quality of the deblurring image, we introduce a measurement, namely Haar defocus score that takes advantage of the Haar-Wavelet transform. The experimental results prove that the proposed method obtains a high quality deblurred image with respect to both the Haar defocus score and the Peak Signal to Noise Ratio.**

## I. Introduction

Image blurring is the result of the light diverting to the other neighborhood pixels rather than falling on the center on the exposure time of imaging. Some common reasons, such as the long exposure, the moving objects, the camera shaking, and the out of focus, lead to an unexpected blurring image.

Considering the motion blur as the shift-invariant through the image, the blurring model can be obtained by adding a noise followed by the convolution of the true sharp image with a blur kernel (point spread function). Image deconvolution can be divided into non-blind and blind methods. The process of recovering the sharp image from a given blurred image and a known kernel is called the non-blind deconvolution [1, 2, 3, 4, 5]. In cases of blind deconvolution, the blur kernel and the sharp image need to be estimated from a single blur image [6, 7, 8]. Both blind and non-blind deconvolution methods are ill-posed problems since the observed data is less than unknown variables [3].

In order to reduce the ill-posedness of blind deconvolution, the regularization terms are proposed to add to the solution. The recent previous works are basically classified into three categories. First, maximum posterior based methods (MAP) [7, 17] aim to seek the most probable solution by maximizing a posterior distribution. This class of methods is simple but facing the problems of convergence. Second, variational Bayesian based methods have been proposed [18], which looks for all possible solutions instead of the most probable solution. The methods in this category are stated to be more robust than the MAP-based methods [18]. However, as a result of seeking all possible solutions, these methods are slow and hard to be applied in real applications. Differ from previous two approaches, edge prediction-based methods [2, 12] are explicitly trying to recover sharp edges by using some heuristic image filtering or iterative thresholding processes. This definitely is an efficient way and has been proved to be effective in practice. In [3], Krishnan *et al.* modelled the gradients in the natural image as a function of hyper-Laplacian to minimize the cost function that corresponds to the true sharp image. Other assumptions such as Gaussian and Laplacian priors were introduced in [4, 5] to recover images from a given blurred image and blur kernel in non-blind deconvolution cases.

Based on the observation that a natural image has lots of edges pointing to arbitrary directions, this paper focuses on analyzing the influence of edge information in the image blind deconvolution process. Instead of using gradient information in only $x$ and $y$ directions as in the previous methods, we use the Gabor filters to obtain richer gradient information for optimizing the cost function. Using the Gabor filters[16] we can control the number of filters for calculating the edge information by setting different values of angle. The Gabor filters allow us to control the number of filters that use to compute the edge information by setting different angle values. This paper also proposes a novel method for evaluating the quality of deblurred image. We rely on the observation that in a blurred image the differences of pixel values are much lower than those in a natural image. We process the Haar-Wavelet transform for those images and define scores for them by using value of standard deviation. These scores are called Haar defocus scores. This is very useful method for evaluating the quality of deblurred results, especially for images without ground truth.

This paper is organized as follow: Section II introduces in detail the concept of the blind deconvolution process. Section

III presents the proposed method with the analysis of applying Gabor filters, and the optimization steps for estimating both the kernel and the latent image are also described. Section IV then gives the definitions of deblurring quality measurement. Section V shows the experimental results and comparison with existing methods. Finally, the conclusion is drawn in Section VI.

## II. IMAGE DECONVOLUTION

As other existing deconvolution methods, this paper relies on assumption that the blur kernel is uniform over entire image. Therefore, the blurred model can be defined in Eq. (1) [7]:

$$y = x \otimes k + N \qquad (1)$$

where $y$ is the observed blurred image, $x$ denotes the true sharp image needed to recover, $k$ is the blur kernel, and $N$ is the random noise.

Our goal is to find the sharp image $x$ from single blurred image $y$ without knowledge of the kernel $k$. This is equivalent to maximize the joint probability of the sharp image $x$ and the kernel $k$ given single blurred the image $y$ or maximize a posterior (MAP) probability:

$$\max_{x,k} p(x,k|y) \propto \max_{x,k} p(y|k,x)\, p(x)\, p(k) \qquad (2)$$

As shown in Eq. 2, to maximize a posterior probability, the prior knowledge of the natural sharp image $x$ and the kernel $k$ should be utilized. The cost function of deblurring process can be written by taking the logarithm of Eq. (2):

$$\mathcal{L} = \min_{x,k} \|x \otimes k - y\|_2^2 + \varphi(x) + \xi(k) \qquad (3)$$

The first term is the fitting term, the second and third terms $\varphi(x)$ and $\xi(k)$ are the regularization terms of the sharp image $x$ and the kernel $k$, respectively. Finding $x$ and $k$ in Eq. 3 becomes to find the optimal solution of the convex function $\mathcal{L}$. There are lots of forms for regularization terms [7] of $x$ such as $||x||_0$, $||x||_1$, $||x||_2$, and $||x||_1/||x||_2$ proposed and used in this field. In this paper we use $||x||_1/||x||_2$ for the sharp image $x$ and $||k||_1$ for the kernel $k$ which are stated to give the lowest value for cost function [3, 7]:

$$\mathcal{L} = \min_{x,k} \sum \left( \alpha \|x \otimes k - y\|_2^2 + \frac{\|x\|_1}{\|x\|_2} + \beta \|k\|_1 \right) \qquad (4)$$

where $\alpha, \beta$ are the factor of regularization terms controlling the strength of the true sharp image and the blur kernel, respectively.

To solve the Eq. 4, the Expectation Maximization (EM) method [19] can be used. The basic idea of EM is to process iteratively in two steps: i) $EM_x$ fixes the $k$ to estimate the $x$, and ii) $EM_k$ fixes the $x$ to estimate the $k$; therefore, the cost function for $EM_x$ and $EM_k$ optimization steps become:

$$\mathcal{L}_x = \min_x \sum \left( \alpha \|x \otimes k - y\|_2^2 + \frac{\|x\|_1}{\|x\|_2} \right) \qquad (5)$$

$$\mathcal{L}_k = \min_k \sum (\zeta \|x \otimes k - y\|_2^2 + \|k\|_1) \qquad (6)$$

where $\zeta = \alpha/\beta$

Fig. 1 shows the entire process of recovering the true sharp image which consists of the kernel estimation and the non-blind deconvolution steps.

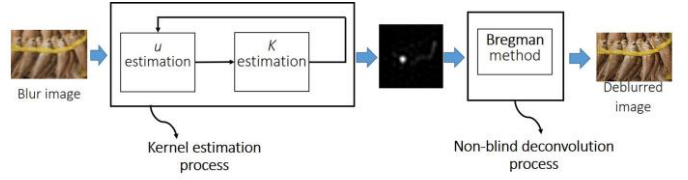

Fig. 1. The process of image deconvolution consisting of kernel estimation and non-blind deconvolution

## III. PROPOSED BLIND DECONVOLUTION

The proposed deconvolution algorithm is summarized in the following pseudo-code:

| **Algorithm**: Overall Algorithm |
| --- |
| **Input:** Observed blurry image $y$, Maximum kernel size ($h \times h$) |
| 1. Blind estimation for kernel $k$ |
|    Create a vector pyramid of kernel size with scale $s$ |
|    Create an image pyramid using scale $s$ with level index $\in \{1,2,\cdots,m\}$ |
|    **For** image level index $i = m$ (smallest level) down-to 1, do |
|      (a) Get the blurry image $y_i$ at the level in image pyramid |
|      (b) Initialize kernel and latent image |
|      (c) Using Gabor filters on $y_i$ to generate high frequency gradient images $g^{y_i}$ |
|      (d) Estimate latent image at the level using $g^{y_i}$ images and FISTA algorithm |
|      (e) Estimate kernel at the level using the estimated latent image and IRLS algorithm |
|      **Upscale** to the next level |
|    **End for** |
|    **Return** blur kernel $k$ |
| 2. Deblurring using non-blind deconvolution algorithm [3] |
|      (e) Deblur the blurry image to obtain the sharp image |
|    **Return** sharp image |

### A. Kernel and latent image initialization

For the smallest level $m$, the kernel is created with the size defined from the level $m$ of kernel size pyramid. The kernel is initialized with random values in the range [0, 1]. Also, the latent image is initialized by applying the Gabor filter bank on the blur image in the image pyramid at level $m$. In the next upscale level, the kernel and latent image are initialized by up-resizing the kernel and the latent image estimated in the previous level by scale $s$. The number of image levels is decided by the kernel size. Namely, since we predefined the minimum and maximum of the kernel size and given the ratio $s$ between two nearby kernel sizes, we can easily determine the number of image levels $m$ because each level will yield a temporary kernel. And we set $s=\sqrt{2}$ to obtain an adaptive number of layers.

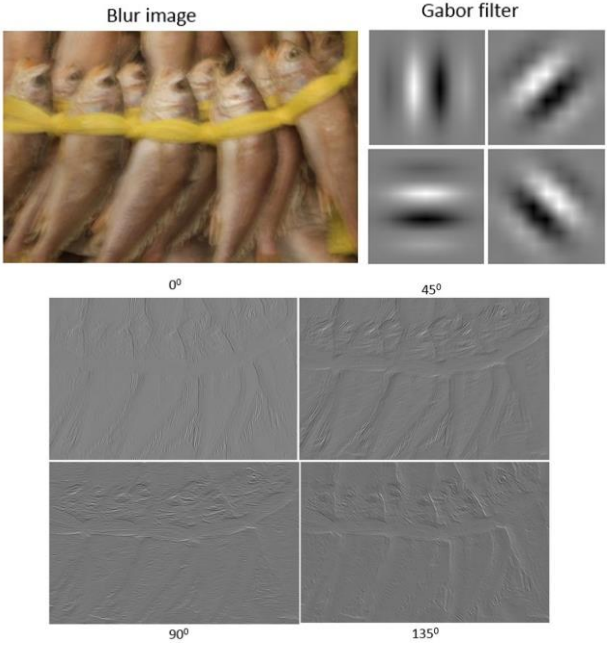

Fig. 2. Gradient information obtained by using different Gabor filters

### B. Gabor filter-based edges extraction

In order to obtain a richer edge information from image, the Gabor function is used to create directional filters. Recall that the Gabor function is the multiplication of Gaussian function and cosine wave:

$$g(u,v;\lambda,\theta,\psi,\sigma,\gamma) = \exp\left(-\frac{u'^2+\gamma^2 v'^2}{2\sigma^2}\right)\cos(2\pi\frac{u'}{\lambda}+\psi) \quad (7)$$
$$u' = u\cos\theta + v\sin\theta \quad v' = -u\sin\theta + v\cos\theta$$

where $u$ and $v$ are the coordinates of image pixel in horizontal and vertical direction, respectively, $\lambda$ represents the wavelength of the cosine factor, $\theta$ denotes the orientation of the normal to the parallel stripes of Gabor function, $\psi$ is the phase of offset, $\sigma$ the standard deviation of the Gaussian function, and $\gamma$ represents the spatial aspect ratio.

In this work, we use $\lambda = 0.5, \psi = 90^0, \sigma = 4, \gamma = 1$, and change the parameter $\theta$ to create 4 directional filters, $\theta = \{0^0, 45^0, 90^0, 135^0\}$. Using this filter bank, the omnidirectional gradient images are built by convolving the input image $y$ with the filter in each direction and then concatenate to obtain $g^y = [g^y_{\theta_1},\dots,g^y_{\theta_n}]$, $n=4$ in our work. Those filters are used to extract gradient information in corresponding directions. Using Gabor function, we can vary the number of filters by changing the value of $\theta$. The influence of $\theta$ to the quality of deblurred image is provided by experiments in Section V. An example of directional gradient extraction is shown in Fig. 2.

### C. Latent image EMx estimation process

The goal of this step is to find the $x$ that minimize cost function $\mathcal{L}_x$ in Eq. 5. We take $\alpha = 100$ as shown to be the optimal value in [7]. By using the regularization term $\|x\|_1/\|x\|_2$, the cost function $\mathcal{L}_x$ becomes non-convex, thus the local and global minimum are not identical [7]. In order to effectively solve Eq. 5, we can use the iterative method and fix $\|x\|_2$ norm to solve the convex optimization problem

$$\mathcal{L}_x = \min_x \sum(\mu\|x \otimes k - y\|_2^2 + \|x\|_1) \quad (8)$$

and its matrix form

$$\mathcal{L}_x = \min_x \sum(\mu\|Kx - y\|_2^2 + \|x\|_1) \quad (9)$$

where $\mu = \alpha\|x\|_2$

Solving convex optimization problem $\|x^i\|_1$ can be done directly. In this paper we use Fast Iterative Shrinkage-Thresholding Algorithm (FISTA) [13] rather than Iterative Shrinkage-Thresholding Algorithm (ISTA) [11] which is used in [7], since the complexity of FISTA $O(1/j^2)$ is faster than the complexity of ISTA $O(1/j)$ where $j$ represents the iteration counter.

| Algorithm: FISTA |
|---|
| **Input:** blur image $y$, concatenated latent image $x_0$, regularization parameter $\mu$, initial kernel $k$, threshold t = 0.001, number of iteration $M = 2$ |
| **Output**: Updated latent image $x$ |
| Step 0: $z_1 = x_0 \quad q_1 = 1$ |
|   Step 1: Generate high frequency images $g$ from $y$ |
|   Loop for j=1 to $M$ |
|     Step 2: $x_j = \mathcal{E}_{\mu t}\left(z_j - tk^T(kz_j - g)\right)$ |
|     Step 3: $q_{j+1} = \frac{1+\sqrt{1+4q_j^2}}{2}$ |
|     Step 4: $z_{j+1} = x_j + \left(\frac{q_j-1}{q_{j+1}}\right)(x_j - x_{j-1})$ |
|   End loop |
| **Return** latent image $x$ |

$\mathcal{E}_{\mu t}$ is soft shrinkage defined as
$$\mathcal{E}_{\mu t}(x_j) = \max(|x_j| - \mu t, 0)sign(x_j) \quad (10)$$

The main difference between ISTA and FISTA is that the iterative shrinkage $\mathcal{E}_{\mu t}(.)$ do not rely on previous image $x_{j-1}$ but image $z_j$ which uses a linear combination of $x_{j-1}$ and $x_{j-2}$.

### D. Kernel EM$_k$ update process

The form for updating kernel is given in Eq. 6 subject to constraints $k \geq 0, \sum_i k = 1$. This optimization can be solved by using Iterative Reweighted Least Squares (IRLS) [12, 14]. In order to make the IRLS converge faster, we also use the Conjugate Gradient (CG) [15] to update the $k$ for each $\omega_i$ updated.

| Algorithm: IRLS |
|---|
| **Input:** concatenated latent image $x$, initial kernel $k$, blur image $y$, parameter $\zeta = 1000$ |
| **Output**: Updated kernel $k$ |
|   Loop for $i=1$ to $N_l$ |

Step 1: $\omega_i = (\zeta k_i)^{-1}$
Loop for $j$=1 to $N_2$
   Step 2: calculate residual for each component in x:
$r_j = x_j^T y_j - (x_j^T x_j k_i^j + \omega_i k_i^j)$
   Step 3: calculate direction for CG: $d_j = x_j^T x_j r_j + \omega_i r_j$
   Step 4: calculate step size: $\mu_1 = \| r_j^T r_j \|_2^2 / \| d_j^T x_j d_j \|_2^2$
   Step 5: update kernel and residual:
     $k_i^{j+1} = k_i^j + \mu_1 d_j, \quad r_{j+1} = r_j - \mu_1 d_j$
   Step 6: calculate step size: $\mu_2 = \| r_{j+1}^T r_{j+1} \|_2^2 / \| r_j^T r_j \|_2^2$
   Step 7: update direction: $d_{j+1} = r_{j+1} + \mu_2 d_j$
  End loop
End loop
**Return** $k$

where $r_j$ is the residual between the observed image and the desired image, $d_j$ represents the direction for updating CG, and $\mu_1$ and $\mu_2$ are step sizes. In the experimental result, $N_1$= 3 and $N_2$ =5 are used.

By combining the CG inside the IRLS, the process of estimating kernel $k$ can be converged faster, in 1 or 2 iterations. With the usage of IRLS, we discovered that the convergence of the entire kernel estimation process strongly depends on the kernel initial distribution. The weight $\omega$ is defined based on residual of the current kernel and the desired one. That means, the larger residual we have, the lower weight we obtain. Therefore, instead of predefining the initial kernel with a prior distribution [7], this paper uses a uniform distribution for the kernel initialization which gives the kernel estimation process converged after 5 iterations rather than 21 as in [7].

*E. Final non-blind deconvolution*

In order to obtain a high quality deblurred image, this paper also uses a coarse-to-fine structure, which is based on image pyramid. In this structure, the estimating process begins from the coarse level (the image lower resolution) to the fine level (the image in higher resolution). The results of kernel and latent image estimation in the previous level are taken as the input parameters in the next level. The final estimated kernel $k$ is the result produced in the finest level.

After having kernel $k$ estimated, the existing non-blind deconvolution methods such as Richardson-Lucy [9], Yuan et al. [10], Fergus et al. [3] etc. can be applied. This paper uses non-blind deconvolution method, proposed by Fergus which is quite robust to the noise and extremely suitable for artifact reduction.

## IV. METHODS OF QUALITY MEASUREMENT

*A. Peak Signal to Noise Ratio (PSNR)*

The quality of deblurred image varies from human to human. Therefore a numerical method for quality measurement is necessary. For doing that, the PSNR can be used to measure the quality of deblurred image when the ground truth is available. The mathematical equation of PSNR is defined as follow:

$$PSNR = 10 log_{10} \frac{MAX_I^2}{MSE} \quad (11)$$

where $MAX_I$ is the maximum possible value of the image, and $MSE$ is the mean square error.

$$MSE = \frac{1}{m.n} \sum_{i=0}^{m-1} \sum_{j=0}^{n-1} [I(i,j) - G(i,j)]^2 \quad (12)$$

where $I(i, j)$ and $G(i, j)$ are pixel value of deblurred and ground truth image, respectively; $m$ and $n$ are the size of the image.

The PSNR measures the ratio between the maximum possible power of a signal and the power of corrupting noise that affects the fidelity of its representation. The larger PSNR value we have, the better deblurred image we obtain.

*B. Score of defocus based on Haar-Wavelet transform*

This paper also proposes a method to calculate the score of deblurred image using the Haar-Wavelet transform when the ground truth is not available. Based on the fact that the gradient information in the true sharp image is much sparser than gradient in the blur image. The defocus score can be defined as:

$$Q_B = e^{-\sigma_D} \in [0,1] \quad (13)$$

where $Q_B$ is the defocus score, $\sigma_D$ represents the standard deviation of the histogram of the diagonal component.

## V. EXPERIMENTAL RESULTS

*A. Real image dataset*

The images used to test the proposed algorithm are those used in the previous works [7, 8] for the aim of comparison. The deblurred images are obtained by using different numbers of Gabor filters. The defocus scores are then calculated to decide how many Gabor filters are most suitable for high quality image deconvolution. The results also compare with some recent advanced existing methods such as Xu [8] and Krishnan [7] based on the defocus score.

As shown in the Table I and Fig. 4, the best quality of deblurred image is produced by using 3 Gabor filter of $60^0$ intervals, $\theta = \{0^0, 60^0, 120^0\}$. In image de-convolution process, creating a lot of filter for edge detection not only worsens the quality of deblurred image, but also increases the processing time. Therefore, the best deblurred image produced by using 3 Gabor filters is selected to compare with recent existing methods such as Xu [8] and Krishnan [7] as shown in Fig. 5. In comparison with other methods, our method gives a shaper deblurred image. This result is continuously confirmed in Table II of the defocus score.

*B. Synthetic image dataset*

We also test the proposed method with synthetic images from the dataset in Levin et al. [20]. When the ground truth is available, the quality of deblurred image can be measured by PSNR. In this paper, we also compare the deblurred results of recent advanced methods such as Xu [8], Yuan [10], and Krishnan [7] using the PSNR are shown in Table III, Fig. 6 and Fig. 7. The result of the proposed method (PSNR=33.57) is significantly better than Xu's method (PNSR=32.47) with the pagoda image. In the church image, result of this paper (PSNR=32.94) is not good as Xu's method (PSNR=33.24). The main reason is, the proposed method relies on the edge information which is not clear in the dim church image.

TABLE I
DEFOCUS SCORE OF SHARP IMAGE PRODUCED USING DIFFERENT NUMBER OF GABOR FILTERS

| Image | Size | Defocus score | | | | | |
|---|---|---|---|---|---|---|---|
| | | #3 | #4 | #5 | #6 | #8 | Blur |
| fishes | 858×558 | **0.1061** | 0.1065 | 0.1063 | 0.1071 | 0.1068 | **0.5682** |
| lyndsey | 1024×1280 | **0.0237** | 0.0352 | 0.0354 | 0.0324 | 0.0341 | **0.1388** |

TABLE II
DEFOCUS SCORE OF SHARP IMAGE USING DIFFERENT METHODS

| Image | Size | Defocus score | | | |
|---|---|---|---|---|---|
| | | Xu [8] | Krishnan [7] | Ours | Blur |
| fishes | 858×558 | 0.1459 | 0.1456 | **0.1061** | 0.5682 |
| lyndsey | 1024×1280 | 0.0478 | 0.0514 | **0.0237** | 0.1388 |

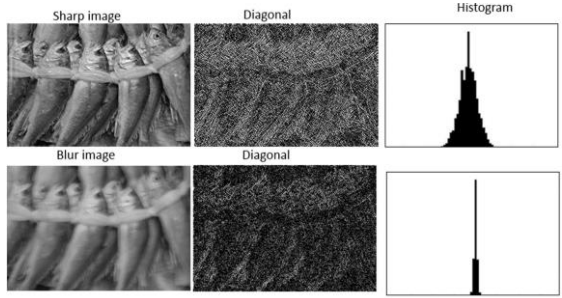

Fig. 3. Defocus score for quality measurement based on the histogram of diagonal image in Haar-Wavelet transform

TABLE III
PSNR OF SHARP IMAGE USING DIFFERENT METHODS

| Image | Size | PSNR | | | | |
|---|---|---|---|---|---|---|
| | | Blur | Xu [8] | Yuan [10] | Krishnan [7] | Ours |
| pagoda | 800×800 | 22.83 | 32.47 | 31.26 | 32.01 | **33.57** |
| church | 800×800 | 27.58 | 33.24 | 31.66 | 32.85 | **32.94** |

## VI. CONCLUSIONS

Taking advantage of a natural image which has a rich omnidirectional edge information, this paper proposes a method of combining the edge information in different directions for deblurring image. Various numbers of Gabor filters are created for analyzing the performance, the results obtained by using 3 Gabor filters give the best score in term of Deblur score and PSNR. However, due to relying on edge information, the proposed method behaves not very well toward the dim image in which the edge is not fully available. The future work will focus on evaluating the proposed method on the whole database for a thorough comparison.

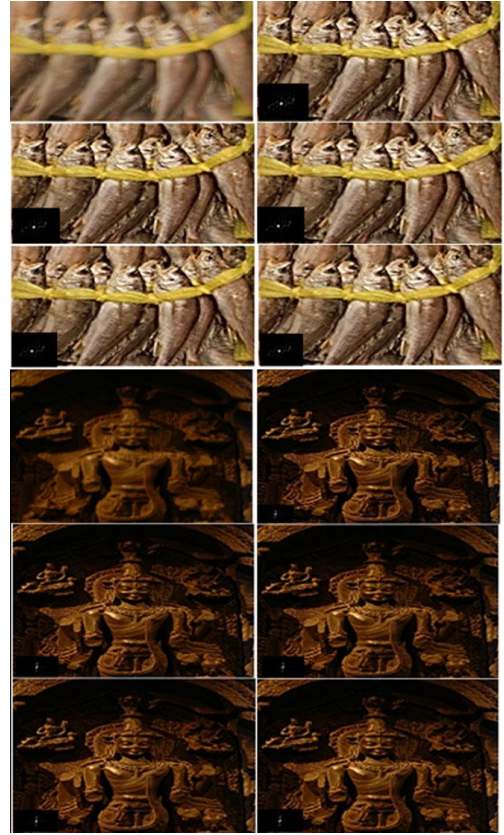

Fig. 4. The deblurred results of fishes and lyndsey images using different number of Gabor filters. First row: left – blur, right –using 3 Gabor filters. Second row: left –4 and right –5 Gabor filters. Last row: left –6 and right –8 Gabor filters.

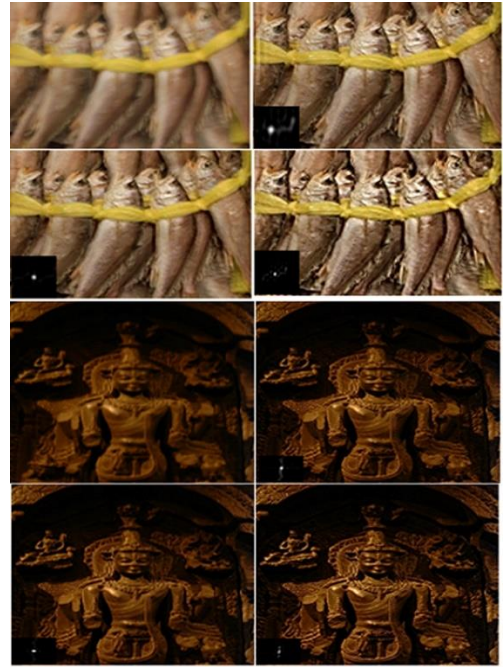

Fig. 5. The deblurred results of fishes and lyndsey images using different methods. First row: left – blur image, right – Krishnan [7]. Second row: left – Xu [8] right – our method

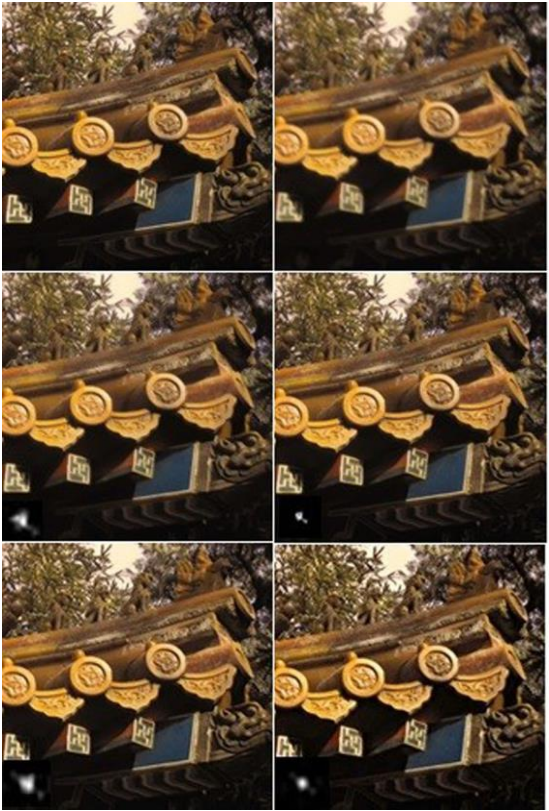
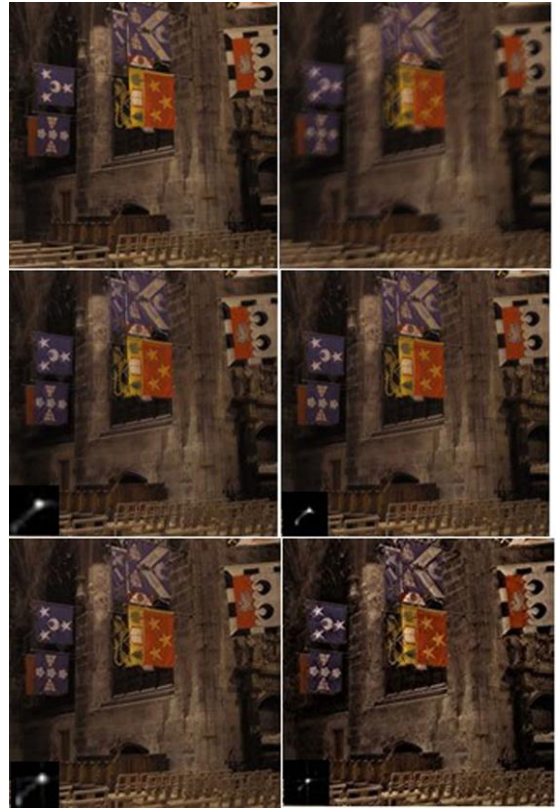

Fig. 6. The deblurred results of pagoda image using different methods. First row: left – ground truth image, right – blur image. Second row: left – Yuan [10], right – Xu [8], last row: left – Krishnan [7], right – our method

Fig. 6. The deblurred results of pagoda image using different methods. First row: left – ground truth image, right – blur image. Second row: left – Yuan [10], right – Xu [8], last row: left – Krishnan [7], right – our method